\documentclass[sn-mathphys]{sn-jnl}
\usepackage{underscore}

\jyear{2021}%

\theoremstyle{thmstyleone}%

\theoremstyle{thmstyletwo}%

\theoremstyle{thmstylethree}%

\begin{document}

\title[Prostate cancer stratification]{Reusability report: Prostate cancer stratification with diverse biologically-informed neural architectures}


\author[1,2,3]{\fnm{Christian} \sur{Pedersen}}\email{c.pedersen@nyu.edu}

\author[4]{\fnm{Tiberiu} \sur{Tesileanu}}

\author[5]{\fnm{Tinghui} \sur{Wu}}

\author[4]{\fnm{Siavash} \sur{Golkar}}

\author[6,7,8,3]{\fnm{Miles} \sur{Cranmer}}

\author*[5,9]{\fnm{Zijun} \sur{Zhang}}
\email{zijun.zhang@cshs.org}

\author*[3,8,2]{\fnm{Shirley} \sur{Ho}}
\email{shirleyho@flatironinstitute.org}

\affil[1]{\orgdiv{Courant Institute of Mathematical Sciences}, \orgname{New York University}, \orgaddress{\city{New York}, \country{USA}}}

\affil[2]{\orgdiv{Center for Data Science}, \orgname{New York University}, \orgaddress{\city{New York}, \country{USA}}}

\affil[3]{\orgdiv{Center for Computational Astrophysics}, \orgname{Flatiron Institute}, \orgaddress{\city{New York}, \country{USA}}}

\affil[4]{\orgdiv{Center for Computational Neuroscience}, \orgname{Flatiron Institute}, \orgaddress{\city{New York}, \country{USA}}}

\affil[5]{\orgdiv{Division of Artificial Intelligence in Medicine, Department of Medicine}, \orgname{Cedars-Sinai Medical Center}, \orgaddress{\city{Los Angeles}, \country{USA}}}

\affil[6]{\orgdiv{Department of Applied Mathematics and Theoretical Physics}, \orgname{University of Cambridge}, \orgaddress{ \city{Cambridge}, \country{United Kingdom}}}

\affil[7]{\orgdiv{Institute of Astronomy}, \orgname{University of Cambridge}, \orgaddress{ \city{Cambridge}, \country{United Kingdom}}}

\affil[8]{\orgdiv{Department of Astrophysical Sciences}, \orgname{Princeton University}, \orgaddress{ \city{Princeton}, \country{USA}}}

\affil[9]{\orgdiv{Center for Computational Biology}, \orgname{Flatiron Institute}, \orgaddress{\city{New York}, \country{USA}}}




\maketitle

\section{Introduction}\label{sec1}
The advancement of high-throughput molecular profiling technologies has enabled genome-scale multi-modality characterization of large cohorts of prostate cancer (PCa) patients \cite{tcga2013, icgc2020}. Leveraging the multi-omic big data, individual genes and pathways have been identified as implicated in lethal castration-resistant prostate cancer (CRPC). However, how the molecular interactions and the combined predictive power of these individual molecular features promote CRPC remain elusive. An integrative and interpretable machine learning framework would hold the promise to elucidate the landscape of molecular factors and their interaction in prostate cancer, and more broadly, in biomedical applications.

Elmarakeby et al.~\cite{Elmarakeby2021} developed P-NET, a customized deep neural network model that incorporated well-established hierarchical biological knowledge from genomic profiles. P-NET integrates patient multi-omic molecular features as input to predict localized vs metastatic PCa. Importantly, the hidden layers in P-NET are heavily sparsified and correspond to biological pathways annotated in Reactome \cite{Jassal2019}, such that the hidden neuron activation values represent the biological pathway's importance. This innovative approach unravels interactive molecular events that determine PCa outcomes to capture the convergent biological interactions between multiple molecular events through complete model interpretability. 
The P-NET model not only outperformed other traditional machine learning models but also facilitated the discovery of potentially crucial genes or pathways.

In this report, we verified the reproducibility of the study conducted by Elmarakeby et al., using both their original codebase, and our own re-implementation using more up-to-date libraries. We quantified the significance of the biological information included in the Reactome pathways at fixed sparsification, by comparing randomly sparsified networks with the Reactome-informed P-NET. We found that at fixed sparsity, P-NET performs significantly better, implying the biological information is crucial to P-NET's superior performance. Furthermore, we explored alternative neural architectures and approaches to incorporating biological information into the networks. We experimented with three types of graph neural networks on the same training data, and investigated the clinical prediction agreement between different models. Our analyses demonstrated that deep neural networks with distinct architectures make incorrect predictions for individual patient that are persistent across different initializations of a specific neural architecture. This suggests that different neural architectures are sensitive to different aspects of the data, an important yet under-explored challenge for clinical prediction tasks.

\section{P-NET is reproducible and robust}\label{sec2}

\begin{figure}
    \centering
    \includegraphics[scale=0.4]{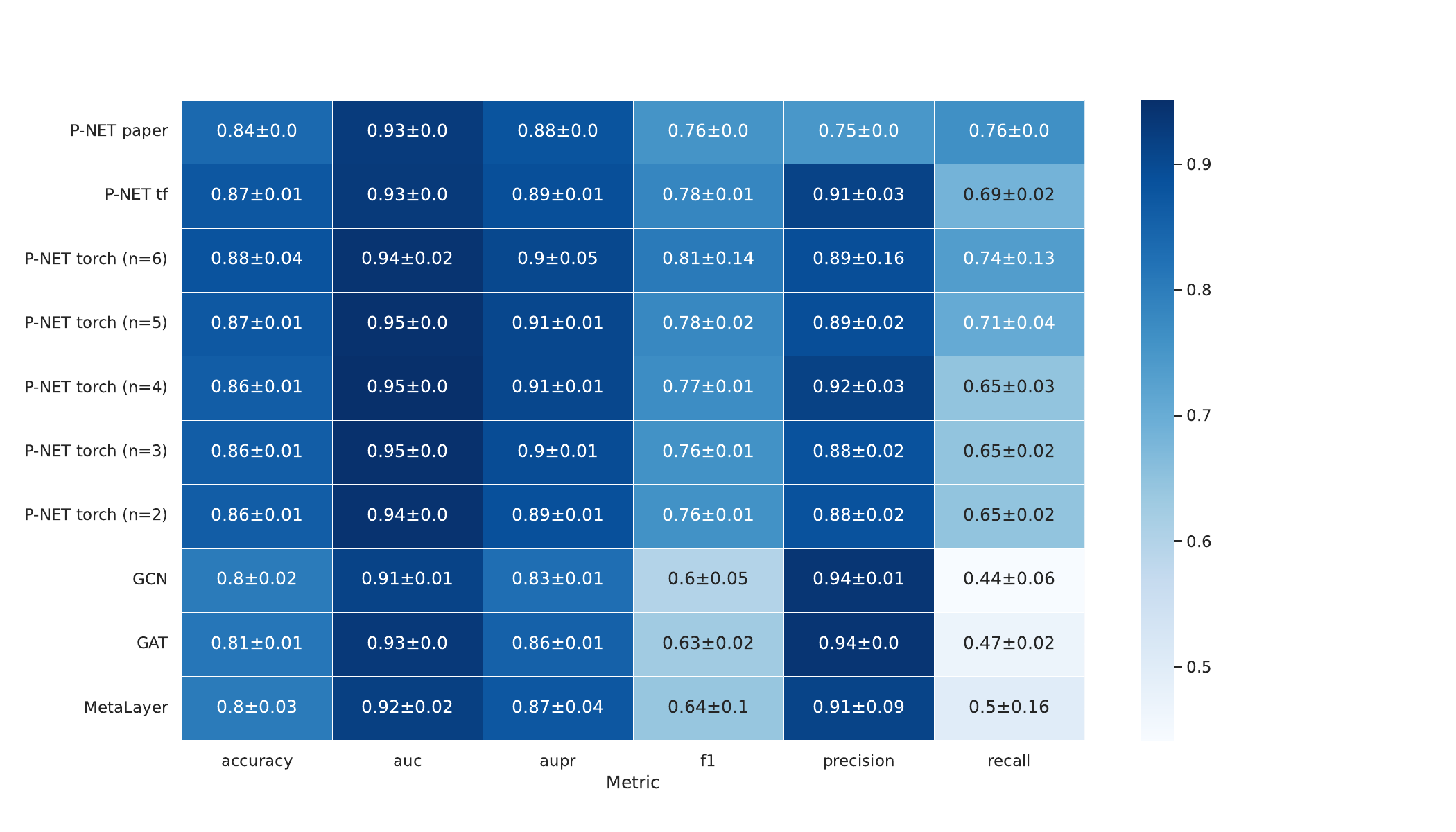}
    \caption{Reproducibility of key metrics on the test set, and comparison for three additional graph networks introduced in section \ref{sec2}. We show the median and variance of 30 models trained using random initialisations. For the P-NET tf results, we use a test set of $n=204$ samples, as in \cite{Elmarakeby2021}. For the P-NET torch and graph networks, half of this set was used for hyperparameter optimisation, and the remaining $n=102$ samples were used to evaluate test performance.}\label{fig:table1}
\end{figure} 

In this section, we attempt to reproduce the results of \cite{Elmarakeby2021}, benchmarking both the original model developed in Tensorflow \cite{Elmarakeby2021} and our refactored implementation in PyTorch. The model presented in \cite{Elmarakeby2021} was evaluated in terms of six key metrics: prediction accuracy, area under the ROC curve (AUC), area under the precision recall curve (AUPR), the F1 score, precision and recall. In Fig.~\ref{fig:table1}, we compare our reproduced values with those provided in the original paper. The top row, ``P-NET paper", shows results taken from Supplementary Table~7 of \cite{Elmarakeby2021}. To produce these results, the model was trained on a training set of $n=807$ samples, and then predictions were recorded on a test set of size $n=204$. The values were obtained by taking the median for each metric after bootstrapping using sample-by-replacement $2,000$ times. The results table of \cite{Elmarakeby2021} did not include variance estimates on these values, which we denote by $\pm0.0$ in our table.

In the second row, ``P-NET tf", we used the publicly available code (https://github.com/marakeby/pnet\_prostate\_paper) released in \cite{Elmarakeby2021} to obtain values for these metrics. We trained the P-NET model with $30$ different random intialisations on the same $n=807$ training dataset. We split the $n=204$ test dataset into two equally sized subsets in accordance with the original codebase of P-NET, one used as a test set, the other will be used as a validation set in the next section. With the exception of batch size (which we set to 10 and was 50 in the original paper), optimization configurations and hyperparameters were kept consistent with the original P-NET code. We tested the $30$ randomly intialised models on the $n=102$ test set, to obtain value for each of the performance metrics in Fig. \ref{fig:table1}, where we show the median and the variance of the $30$ random intialisations.

We subsequently refactored the P-NET codebase in PyTorch \cite{PyTorch}, one of the most popular deep learning frameworks, for better reusability in developing biologically-informed neural networks. This is because the original codebase is written in Python 2.7, using TensorFlow 1 \cite{TensorFlow} for the neural network component, which has been deprecated since October 2021. We re-implemented the P-NET model in Python 3.9 using PyTorch \cite{PyTorch}, and repeated the training and testing procedure as in the TensorFlow case, using the same training and test sets for $30$ initialisations. We find good agreement between the original TensorFlow implementation and our PyTorch implementation, with the biggest differences in the variance across initialisations. We verified that a forward pass of the PyTorch implementation produces an identical output to the TensorFlow model when both architectures have the same weights, confirming that the architectures are consistent. Therefore the small discrepancies in \ref{fig:table1} are likely due to some residual difference in the weight initialisation and optimisation procedures. Additionally, we use a smaller batch size of 10 (where Elmarakeby et al. used 50), and do not use class weights to account for the imbalanced dataset, as was done in the original paper.

The interpretable hidden layers in P-NET represent different granularity levels of biological pathways, controlling for different network complexities and biological interpretation resolution. To explore the effect of incorporating more general, higher-level biological pathways in P-NET and the robustness of P-NET against different interpretable hidden layers, we reduce the number of hidden layers in the network for our PyTorch implementations, from $6$ as in the original architecture, to $2$, and show results for each modified model. Our results demonstrate that the P-NET performance is robust with varying numbers of interpretable hidden layers that represent different abstraction levels of biological pathways. The remaining 3 rows will be discussed in the next section.

\begin{figure}
    \centering
    \includegraphics[scale=0.5]{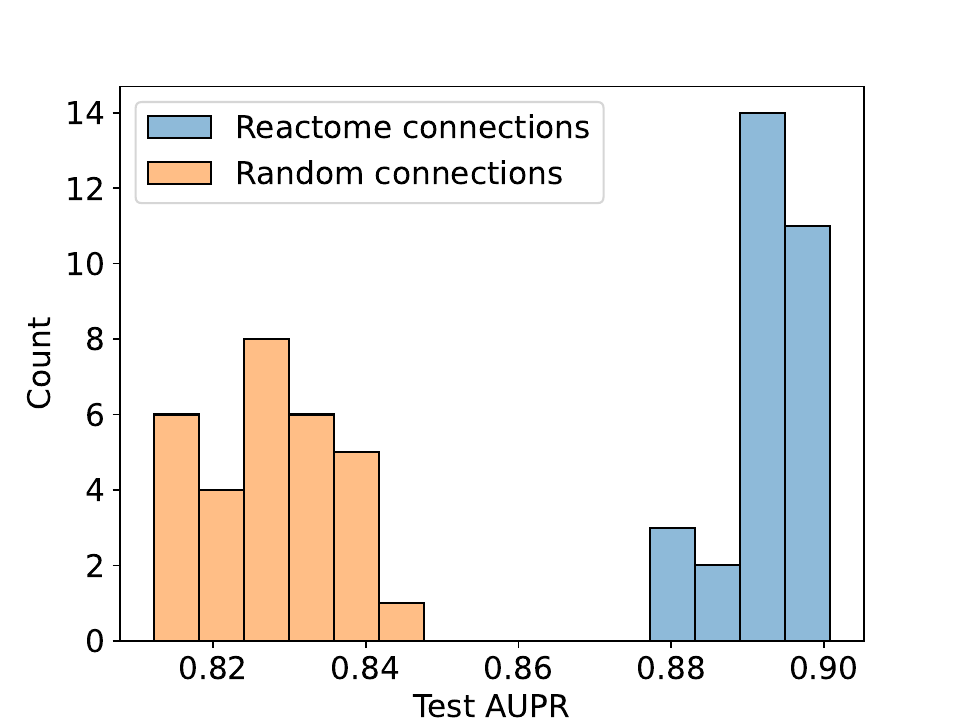}
    \caption{Test AUPR for 30 randomly initialised P-NET models, when using Reactome connections (blue) random connections (orange).}\label{fig:connections}
\end{figure} 

Finally, we investigated to what extent the biological information encoded in the P-NET pathways contributes to model predictions. The P-NET model is heavily sparsified by masking the majority of network connections, except those corresponding to known biological pathways from the Reactome network~\cite{Jassal2019}. We randomly permuted these masks, preserving the number of learnable parameters and the sparsity of the network, but removing the biological information. We trained 30 random initialised models using both the true P-NET connections, and randomised connections, and show the test AUPR in Fig.~\ref{fig:connections}. We observed a significant decease in performance when using randomly permuted connections, implying that the biological information encoded in the Reactome pathways is playing an important role in the accuracy of model predictions.

\section{Graph networks integrate multi-omic features via different biological knowledge}\label{sec3}

In addition to ablating and expanding the original results from \cite{Elmarakeby2021}, we explore alternative architectures based around Graph Neural Networks (GNNs) \cite{Zhou2020}. A fundamental challenge in classifying outcomes based on genetic data is the very high dimensionality of the input and the small number of samples. Because of this, it is crucial to employ a neural architecture with a suitable inductive bias and which incorporates all relevant available domain knowledge. The original work~\cite{Elmarakeby2021}, used this prior information to aggressively sparsify the feedforward neural network, and reduce the number of free parameters. Alternatively, the dimensionality of the problem can be greatly reduced if we can treat all genes equally, that is, if we assume that the classification is invariant under the permutation of the genes' mutation and copy numbers. This assumption does not hold, in particular because the relationship between different genes and their relevance for the occurance of cancer in a specific tissue is not invariant under permutation. However, if we include this information, i.e. the connectivity pattern and the relevance in the structure of a graph, then we can expect that the classification would be approximately invariant under the combined gene and gene connectivity graph permutation.

Following this argument, in order to proceed with GNNs, we need gene importance and gene connectivity pattern information, for which we use the tissue specific network~\cite{Greene2015}. These networks have an interaction strength between genes in the range $[0,1]$. We construct a graph with each gene as a node, and have an undirected, unweighted edge connection between all gene pairs with interaction strength $>0.5$. We consider three methods of graph deep learning. The simplest is a graph convolutional network (GCN). Additionally, we apply a Graph Attention Network (GAT) \cite{Velickovic2017}, and MetaLayer \cite{Battaglia2018} to the same datasets. For each graph network, we optimise hyperparameters in the following way. For all three networks, we vary the learning rate in the range $[0.1,0.00001]$, where the samples are drawn from a log uniform distribution. The number of layers (or message passing steps) in each graph model is drawn from the range $[2,6]$. For the GCN, the size of the latent node representations and the size of the final fully connected layer are independently drawn from $[32,64,128,256]$. For the GAT, we use the same range for the latent node representations as for the GCN, and include an additional hyperparameter ranging from $[1,8]$ to describe the number of self-attention heads. In the case of multi-head attention networks, we combine the attention scores through averaging instead of concatenation, to prevent the size of the latent features becoming too large in the case of networks with several layers. Finally, for the MetaLayer, we use two-layer perceptrons for the node and edge model, and vary one hyperparameter describing the size of these hidden layers in range $[32,64,128,256]$.

We draw 50 random samples from each graph's hyperparameter space, and train each model on the same training set described in the previous section. Recall that the original paper had a test set size of $n=204$, and our results in Fig. \ref{fig:table1} used a test set of size $n=102$. We treat the remaining $n=102$ samples as a validation set, and evaluate the performance of each draw of hyperparameters on this validation set. We select the network which has the highest AUPR on the validation set as our best performing network. To obtain evaluate the test set performance, we repeat the testing procedure from the previous section, where each model is initialised $30$ different times and trained on the training set, and test metrics are estimated as the median and standard deviation over these $30$ samples. This process is repeated for each of the three graph network models. We observed slightly worse performances of GNNs compared to P-Net (Fig. \ref{fig:table1}). We delve into the detailed differences between the model predictions in the next section.

\begin{figure}
    \centering
    \includegraphics[scale=0.65]{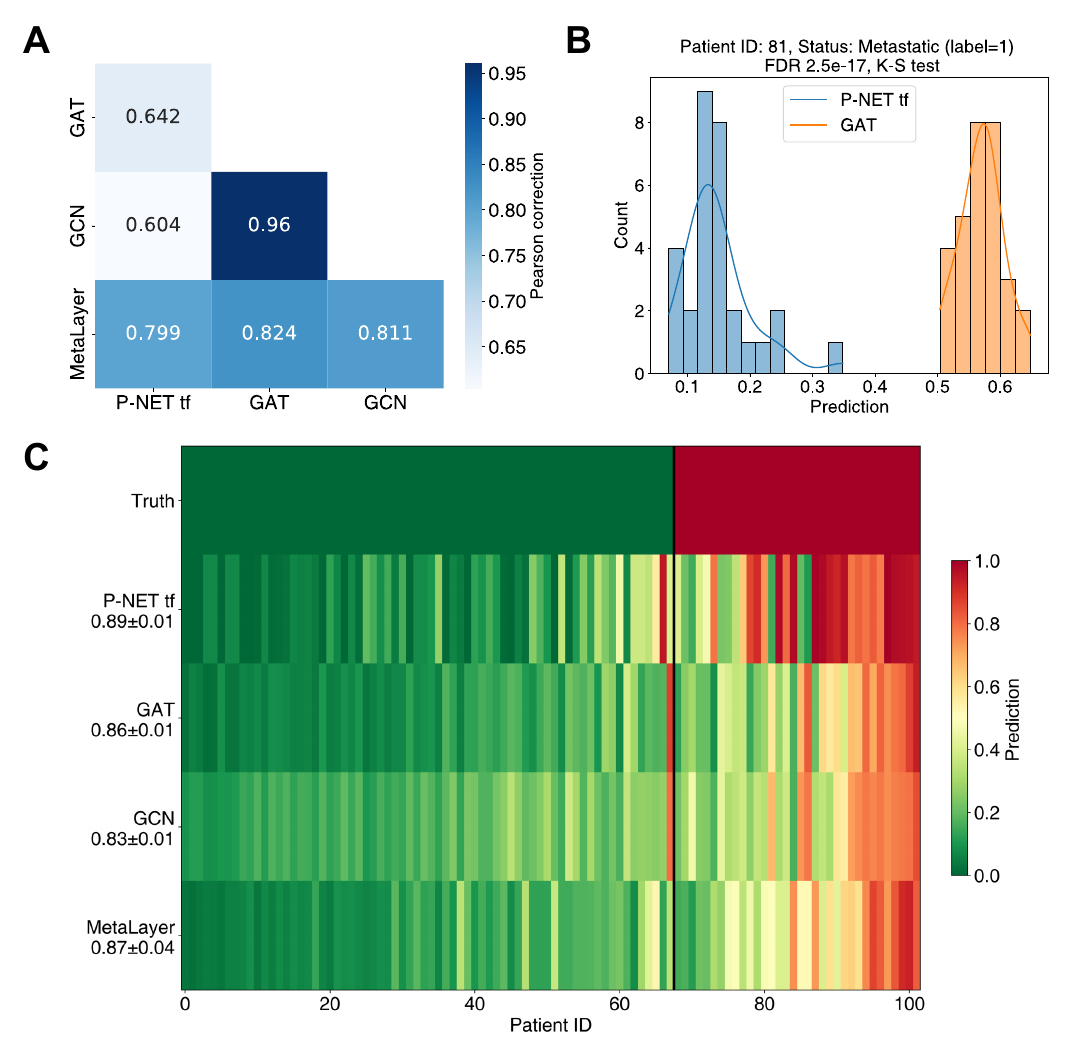}
    \caption{A: Correlation matrix between predictions of different neural architectures, when trained and tested on the same dataset. B: Distribution of prediction over 30 initialization on one positive test case. C: Comparison of model predictions on test set for different neural architectures. The numerical value under each architecture label shows the median AUPR on the test set, when training 30 randomly initialised models.}\label{fig:merge}
\end{figure} 

\section{Impact of AI design on per-patient prediction}\label{sec4}
We systematically examined the impact of neural network architectures for P-NET, the fully-interpretable neural network and the various GNNs, on a per-patient basis. This investigation is significant, because the clinical predictions made by such models have profound implications for each patient. Whilst both P-NET and the graph networks introduced in section \ref{sec3} heavily leverage domain knowledge, both the information used (Reactome \cite{Jassal2019} vs HumanBase \cite{Greene2015}) and the way in which this information is incorporated into the model (MLP vs GNN) are different. Therefore, despite all models being trained using the same training data, we hypothesized they may have distinct predictive biases.

When correlating the test-set predicted probabilities between each method, we found P-NET had a relatively low correlation with the GNNs (average PCC=0.682), compared to PCC=0.960 between GCN and GAT (Fig. \ref{fig:merge}A). On an individual patient basis, there exist metastatic PCa patients that were predicted correctly by GAT but were misclassified by P-NET (Fig. \ref{fig:merge}B), despite P-NET having an overall better prediction accuracy (AUPR P-NET=0.89$\pm$0.01, GAT=0.86$\pm$0.01). Such prediction differences and errors cannot be explained by random weight initialisation and are only attributable to different neural network designs and inductive biases. Across all test-set patients (n=102), each model made different predictions, and consequently, different errors (Fig. \ref{fig:merge}C). Indeed, we found that the PCa metastatic probability predictions for 50 out of the 102 patients differed by at least 10\% between GAT and P-NET (K-S test, FDR$<$0.05), despite their overall similar performance (Fig. \ref{fig:merge}C). 

In summary, we report the in-depth AI design impact analysis for PCa outcome predictions on a per-patient basis. We find the greatest disparity between models with vastly different design (i.e. between GNNs and P-NET). Deep neural networks with distinct architectures can make incorrect predictions for individual patient that are persistent across different initializations of a specific neural architecture. Further studies are required to elucidate architecture-specific incorrect predictions and how to overcome such pitfalls.

\section{Discussion}\label{sec5}
In this study, we investigated the reproducibility of the results of \cite{Elmarakeby2021}. Using the code released with the original paper, we were able to produce results consistent with the published values. We implemented the biologically-informed neural architecture from \cite{Elmarakeby2021} in an up-to-date version of Python and PyTorch, facilitating the reproducibility of P-NET results, and showed that our new implementation is in good agreement with the original code. We investigated the contribution of the biological information contained in the Reactome pathways on model performance, by randomly permuting the pathways, preserving the model sparsity but removing the biologically informed inductive bias. We observed a significant drop in performance of the randomly permuted networks, demonstrating that the biological information plays an important role in P-NET performance. Additionally, we explored alternative neural architectures informed by different biology knowledge, by applying three graph neural networks to the same training dataset. We found slightly reduced performance from the graph neural network architectures, but found relatively low correlation in the model predictions between the graph networks and P-NET models (Fig. \ref{fig:merge}) . This implies that the different neural architectures are sensitive to different aspects of the data, an important yet under-explored challenge for clinical prediction tasks.

\section{Code Availability}
The original P-NET code is available at https://github.com/marakeby/pnet_prostate_paper. Our codebase with PyTorch implementation and GNNs is available at https://github.com/zhanglab-aim/cancer-net.

\bibliography{sn-bibliography}

\end{document}